\documentclass[
]{ceurart}
\usepackage[utf8]{inputenc}
\usepackage{hyperref}
\usepackage[utf8]{inputenc}
\usepackage[T1]{fontenc}

\UseRawInputEncoding

\usepackage{listings}
\begin{document}

\copyrightyear{2024}
\copyrightclause{Copyright for this paper by its authors.
  Use permitted under Creative Commons License Attribution 4.0
  International (CC BY 4.0).}

\conference{CLEF 2024: Conference and Labs of the Evaluation Forum, September 09–12, 2024, Grenoble, France}

\title{DS@BioMed at ImageCLEFmedical Caption 2024: Enhanced Attention Mechanisms in Medical Caption Generation through Concept Detection Integration}

\tnotemark[1]
%% \tnotetext[1]{You can use this document as the template for preparing your publication. We recommend using the latest version of the ceurart style.}

%% \title[mode=sub]{Notebook for the <Lab name> Lab at CLEF 2024}

\author[1,2]{Nhi N.Y Nguyen}[%
email=21521231@gm.uit.edu.vn,
]

\author[1,2]{Huy L.Tu}[%
email=21522173@gm.uit.edu.vn,
]

\author[1,2]{Phuong D.Nguyen}[%
email=21520091@gm.uit.edu.vn,
]

\author[1,2]{Tan N.Do}[%
email=21522575@gm.uit.edu.vn,
]

\author[3]{Triet M.Thai}[%
email=triettm@oucru.org ,
]

\author[1,2]{Thien B. Nguyen-Tat}[%
email= thienntb@uit.edu.vn ,
]
\cormark[1]
\address[1]{Unviversity of Information Technology, Ho Chi Minh City, Vietnam}
\address[2]{Vietnam National University, Ho Chi Minh City, Vietnam}
\address[3]{Oxford University Clinical Research Unit, Ho Chi Minh City, Vietnam}

\cortext[1]{Corresponding author.}

\begin{abstract}
 \textit{Purpose:} Our study presents an enhanced approach to medical image caption generation by integrating concept detection into attention mechanisms. \\
Method: This method utilizes sophisticated models to identify critical concepts within medical images, which are then refined and incorporated into the caption generation process. \\
Results: Our concept detection task, which employed the Swin-V2 model, achieved an F1 score of 0.58944 on the validation set and 0.61998 on the private test set, securing the third position. For the caption prediction task, our BEiT+BioBart model, enhanced with concept integration and post-processing techniques, attained a BERTScore of 0.60589 on the validation set and 0.5794 on the private test set, placing ninth. \\
Conclusion: These results underscore the efficacy of concept-aware algorithms in generating precise and contextually appropriate medical descriptions. The findings demonstrate that our approach significantly improves the quality of medical image captions, highlighting its potential to enhance medical image interpretation and documentation, thereby contributing to improved healthcare outcomes.
\end{abstract}

\begin{keywords}
Medical Caption Generation \sep
Multimodal Learning \sep
Concept Detection \sep
ImageCLEF 2024 
\end{keywords}

\maketitle

\section{Introduction}
The rapid growth of deep learning techniques has profoundly influenced various sectors, notably medical imaging \cite{intro0}. Among these advancements, using neural networks in radiology has garnered significant attention due to its potential to enhance diagnostic accuracy and efficiency \cite{intro0.1}. A particularly intriguing development in this field is the automatic generation of medical captions from radiology images \cite{intro1}. This innovation aims to assist radiologists by providing preliminary interpretations and streamlining clinical documentation. Medical caption generation transforms visual information from radiological images into coherent, clinically valuable language descriptions. This process is inherently challenging due to the complexity and diversity of medical images, the need for precise and context-aware descriptions, and the necessity to incorporate domain-specific knowledge \cite{intro1} \cite{intro2} \cite{intro3}.

Traditional systems often fall short of these requirements, leading to the development of advanced attention mechanisms that can more effectively capture and interpret the intricate details found in radiological images. Recent research shows that integrating concept detection into caption generation algorithms improves performance. Concept detection involves identifying and categorizing critical visual elements in an image, such as anatomical structures, pathological findings, and medical devices. By incorporating these detected concepts into the caption generation process, models can produce more accurate and contextually relevant descriptions. This integration not only enhances the interpretability of the generated captions but also better aligns with radiologists' diagnostic reasoning.
A significant milestone in this area is highlighted by the ImageCLEF campaign, an annual multimodal machine learning competition established in 2003. ImageCLEF fosters advancements in multimedia processing, including computer vision, image analysis, classification, and retrieval in multilingual and multimodal contexts. One of its primary tasks is ImageCLEFMedical, which encompasses challenges such as image annotation, synthetic image creation, and medical image question answering. In ImageCLEF 2024 \cite{ImageCLEF2024}, participants engaged in the ImageCLEFMedical Caption task \cite{ImageCLEFmedicalCaptionOverview2024}, which included two subtasks: concept detection and caption prediction.
Concept detection aims to associate biomedical images with relevant medical concepts, thereby enhancing diagnostic notes by identifying key concepts that should be included in preliminary reports. Moreover, it facilitates the efficient organization and retrieval of medical images by indexing them according to related concepts. Caption prediction, or diagnostic captioning, remains a complex research challenge intended to support the diagnostic process by providing preliminary reports, rather than replacing physicians. This approach aids experienced clinicians in managing high volumes of daily medical examinations more swiftly and efficiently, while also reducing the likelihood of clinical errors among less experienced clinicians.

Our findings demonstrate that incorporating concept detection improves the performance of attention mechanisms and generates more logical and diagnostically valuable captions. This work significantly contributes to the development of intelligent technologies that assist radiologists in their clinical practice, thereby enhancing the standard of patient care. The following sections offer a comprehensive review of relevant literature, detail our proposed methodology, present experimental results, and discuss the implications of our findings. Our goal is to make a substantial contribution to the fields of medical imaging and natural language processing by advancing the capabilities of medical caption generation, thus paving the way for further advancements in automated reporting and medical data interpretation.

\section{Background and Related Works}
\subsection{Former Medical Datasets}
Medical imaging has been a focal point in the application of deep learning, benefiting significantly from the availability of comprehensive datasets. Early datasets such as the NIH ChestX-ray14 \cite{re1} provided a large collection of chest radiographs annotated with disease labels, facilitating advancements in image classification and disease detection tasks. The MIMIC-CXR dataset \cite{re2}, developed by Johnson et al., further enriched the field by offering not only radiographic images but also paired radiology reports, enabling research in image-to-text generation. These datasets have been pivotal in training and validating deep learning models, providing the groundwork for more sophisticated tasks such as medical caption generation and concept detection.

\subsection{Related Work Methods: Concept Detection} 
Concept detection in medical imaging involves identifying and categorizing essential visual elements such as anatomical structures, pathological findings, and medical devices. This task is crucial for generating accurate and contextually relevant medical captions. Early methods primarily relied on traditional machine learning techniques, which often struggled with the complexity and variability of medical images. However, recent advancements in deep learning, particularly convolutional neural networks (CNNs), have significantly improved the accuracy of concept detection. Notable CNN architectures such as ResNet50 \cite{re21} and EfficientNet \cite{re22} have demonstrated substantial improvements in detecting and classifying visual elements in medical images.

Recently, Transformer-based models have been increasingly applied to concept detection due to their ability to capture long-range dependencies and contextual information. Notable examples include Vision Transformer (ViT) \cite{re23}, Bidirectional Encoder representation from Image Transformers (BEiT) \cite{re24}, and Swin Transformer \cite{re25}. These models provide robust feature representations and have shown promise in enhancing the accuracy and interpretability of medical image analysis.

\subsection{Related Work Methods: Caption Prediction}
Caption prediction, or diagnostic captioning, involves generating descriptive text that accurately summarizes the medical content of an image. This task extends beyond simple image annotation, requiring models to produce coherent and clinically meaningful narratives. Traditional captioning methods often used template-based approaches, which lacked flexibility and adaptability to different medical contexts. With the advent of deep learning, particularly sequence-to-sequence models and attention mechanisms, more sophisticated captioning systems have been developed.

For example, Jing et al. proposed a hierarchical LSTM \cite{lstm} model combined with a co-attention mechanism to generate detailed radiology reports from medical images. Their model effectively captured the hierarchical structure of medical reports, producing more detailed and contextually appropriate captions \cite{intro1}. 

The introduction of Transformer models specifically designed for the medical domain has significantly advanced the field of medical image captioning. Transformers, particularly model like BioBERT (Bidirectional Encoder Representations from Transformers for Biomedical Text Mining) \cite{re32}, have demonstrated exceptional capabilities in understanding and generating biomedical text due to their ability to handle complex medical terminology and contexts. Recent research has leveraged these models to improve medical captioning. Additionally, large language models (LLMs) such as BioGPT \cite{re34} have been explored for their potential to generate coherent and diagnostically valuable medical captions, further pushing the boundaries of automated reporting in radiology.

\section{Task and Dataset Descriptions}

\subsection{Task Descriptions}
ImageCLEF has included medical tasks annually since 2004. Since 2019, it has focused each medical task on a specific issue but combined them into a single task with multiple subtasks. Four tasks are proposed for 2024:
Image Captioning, Image Question Answering for Colonoscopy Images, MEDIQA-MAGIC, Quality Control of Synthesized Medical Images Generated by GANs.
In ImageCLEF 2024 \cite{ImageCLEF2024}, we engage in the Image Captioning task \cite{ImageCLEFmedicalCaptionOverview2024}, simultaneously participating in two subtasks: Concept Detection Task and Caption Prediction Task, each crucial in the holistic process of generating informative captions for medical images.
\begin{itemize}
    \item \textbf{Concept Detection Task:} The Concept Detection Task involves using a refined subset of the UMLS 2022 AB version for concept generation. This subset is carefully selected to enhance the accuracy of concept detection by filtering concepts based on their semantic types. Moreover, to optimize concept detection from images, a stringent exclusion criterion is applied to remove low-frequency concepts, based on insights from previous iterations.

    \item \textbf{Caption Prediction Task:} In the Caption Prediction Task, a series of meticulous preprocessing steps are undertaken to ensure the integrity and coherence of the captioning process. Specifically, the removal of embedded hyperlinks within captions is performed as a fundamental preprocessing step. This careful action helps maintain data cleanliness and consistency, thereby supporting subsequent analytical processes and enabling accurate caption prediction outcomes.
\end{itemize}

\subsection{Dataset Information}
The data for the captioning task will consist of images selected from medical literature, including annotations and related UMLS terms manually curated as metadata. A more diverse dataset will be provided to encourage more sophisticated approaches.
For the development dataset, Radiology Objects in COntext Version 2 (ROCOv2) \cite{rocov2}, an updated and expanded version of the Radiology Objects in COntext (ROCO) dataset \cite{roco}, is used for both subtasks. As in previous versions, this dataset originates from biomedical articles in the PMC OpenAccess collection \cite{pmc}, with the test set comprising a set of previously unseen images.
\begin{itemize}
    \item Training Dataset: Includes 70,108 images.
\item  Validation Dataset: Includes 9,972 images.
\item  Test Dataset: Includes 17,237 images.
\end{itemize}

\section{Experiments and Results}

\subsection{The Proposed Approach}
\subsubsection{Concept Detection Methodology}
We aim to extract features from images by carefully examining and testing a variety of pretrained models that fall into three main architectural paradigms, which are shown in Table~\ref{tab:concept}. The list that follows summarizes the particular models that are being examined: 
\begin{itemize}
    \item \textbf{CNN-based architectures:} Microsoft/ResNet-50 \cite{he2016deep}, an archetype of conventional convolutional neural network (CNN) models, characterized by its utilization of residual blocks to mitigate the challenges associated with gradient vanishing, thereby enhancing model performance within computationally tractable bounds.
    \item \textbf{Transformer-based architectures:}
    \begin{itemize}
        \item ViT (Vision Transformer) \cite{dosovitskiy2021image}: Pioneering the paradigm shift in image data processing, ViT adopts a transformative approach by encoding images into patch embeddings, followed by feature extraction using a Transformer encoder, reminiscent of text data processing methodologies.
        \item DeiT (Data-efficient Image Transformers) \cite{touvron2021training}: An evolution of ViT, DeiT emphasizes data efficiency, facilitating training with reduced data volumes while preserving commendable performance metrics.
        \item Swin-v2 (Shifted Window Transformer v2) \cite{swinv2}: Distinguished by its innovative utilization of self-attention mechanisms within shifted windows, Swin-v2 ameliorates computational complexity and augments performance across a spectrum of tasks, including image classification and segmentation.
        \item  BEiT (Bidirectional Encoder representation from Image Transformers) \cite{beit}: At the confluence of Transformer and BERT architectures, BEiT excels in capturing robust image features through bidirectional encoding methodologies.
        \item BiomedCLIP \cite{biomedclip}: A domain-specific adaptation of ViT tailored for biomedical applications, leveraging the CLIP architecture to enhance performance in medical domain tasks.
    \end{itemize}
    \item \textbf{Model Ensembles:}
    \begin{itemize}
        \item Ensemble-2 model (Swin-V2 + BEiT): The symbiotic fusion of Swin-v2 and BEiT engenders a collaborative synergy, capitalizing on the distinctive strengths of each constituent model to surpass individual model performances.
        
        \item Ensemble-4 model (Swin-V2 + BEiT + DeiT + ViT): Comprising a composite quartet of models, this ensemble fortifies accuracy and generalization capabilities through the combination of representatives from Transformer-based models.
    \end{itemize}
\end{itemize}

Following the feature extraction step, the retrieved features pass via a linear layer and classifier, where they are transformed and classified to provide outputs that correspond to the chosen class categories. This key step emphasizes the thorough orchestration of feature transformation and classification to produce predictions specific to the required class taxonomy.

% \usepackage{booktabs}
\begin{table}[h]
\caption{Statistics of models for the Concept Detection subtask
}
\label{tab:concept}
\begin{tabular}{@{}lclc@{}}
\toprule
Models     & Version & Detailed                                      & \# Parameters \\ \midrule
Resnet-50  & -       & microsoft/resnet-50                           & 27122124      \\
BEiT       & base    & microsoft/beit-base-patch16-224               & 88065356      \\
Swin       & v2      & microsoft/swinv2-base-patch4-window12-192-22k & 89459332      \\
DeiT       & base    & facebook/deit-base-patch16-224                & 88692620      \\
ViT        & base    & google/vit-base-patch16-224                   & 88692620      \\
BiomedCLIP & base    & ikim-uk-essen/BiomedCLIP\_ViT\_patch16\_224   & 88692620      \\
BEIT       & large   & microsoft/beit-large-patch16-224              & 305971084     \\
Ensemble   & -       & Swin-v2 + BEiT                                & -             \\
Ensemble   & -       & Swin-v2 + BEiT + DeiT + ViT                   & -             \\ \bottomrule
\end{tabular}
\end{table}

\paragraph{Concept Filtering}
A certain process must be followed while using the BEiT (Bidirectional Encoder Representations from Image Transformers) model in order to carry out idea filtering and modify the output threshold to detect variations in the outcomes. The following are the steps to follow: On a given dataset, do inference using the BEiT model and modify the output threshold to filter the ideas or classes. Setting various threshold values and watching the ensuing outcomes allows for this modification. We may adjust and assess how different thresholds affect the model's performance using this procedure.

\subsubsection{Captioning Methodology}
\begin{figure}[h]
        \centering
        \includegraphics[width=0.78\linewidth]{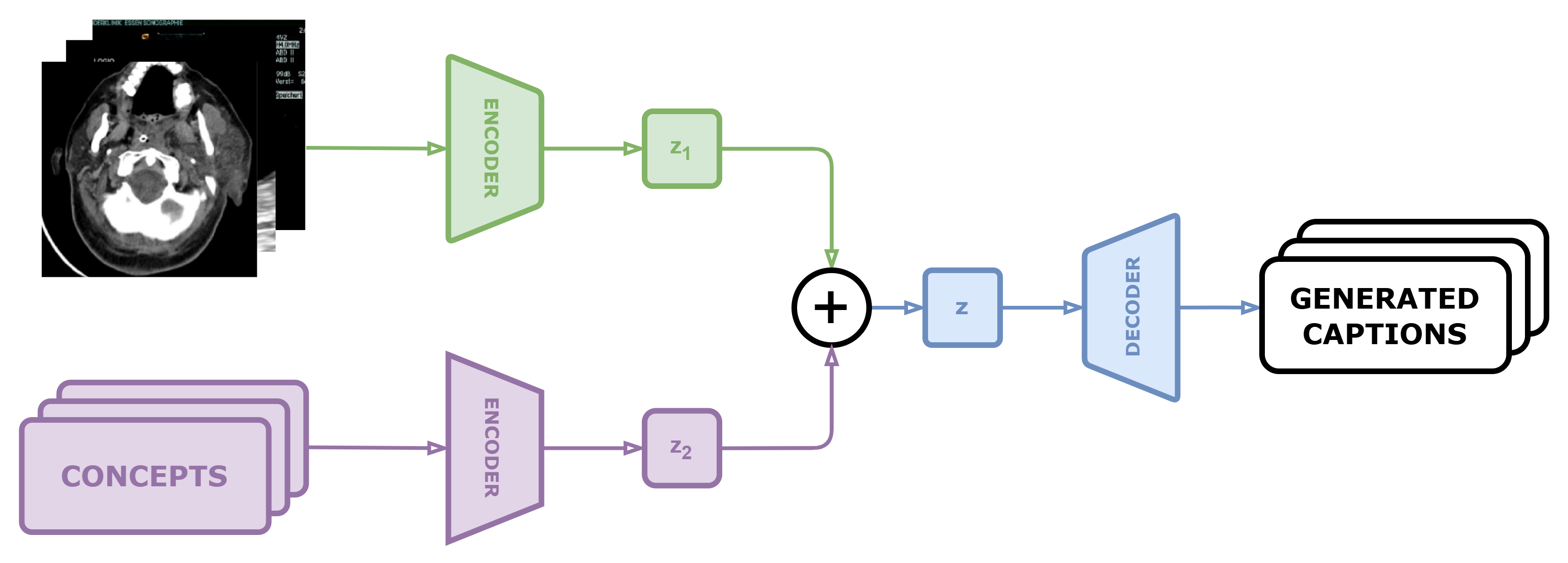}
        \caption{An overview of the multimodal architecture for Medical Caption Generation challenge}
        \label{fig:enter-label_}

\end{figure}

Figure~\ref{fig:enter-label_} depicts an overview of the proposed method for
Medical Captioning task.

Given the primary focus on Image Captioning in this research, the architectural design must effectively extract salient features from both the image and its corresponding text, combining them to generate the final caption. Our carefully curated multimodal fusion architecture incorporates essential components like an image encoder for pertinent feature extraction, a text encoder for eliciting semantic information from text, and a decoder to synthesize insights from the textual context. Additionally, the fusion mechanism integrates image features and output classifications from concept detection, synergistically blending them with textual input to decode and generate the caption output.
The proposed approach leverages the pretrained Bidirectional Encoder Representations from Transformers (BEiT) model for image feature extraction. Boasting a symmetric Transformer architecture, BEiT can comprehend image representational features by concurrently considering both surrounding image patches and global context. With its extensive training on copious data, BEiT can be fine-tuned and achieve state-of-the-art results across several computer vision and image processing benchmarks.

To encode the input text captions, this research employs two domain-specific language models: BioBART (Bidirectional and Auto-Regressive Transformers for Biomedical Text) and ClinicalT5 (Text-to-Text Transfer Transformer fine-tuned on clinical data).
\begin{itemize}
    \item BioBART \cite{BioBART} is a version of the BART model \cite{bart} adapted and further pre-trained on biomedical text data such as medical literature, case reports, and genomic analysis documents. Leveraging its bidirectional Transformer architecture, BioBART can effectively encode both general and biomedical domain-specific text, enabling the extraction of rich semantic representations for tasks like text summarization, medical question-answering, and report generation.
    \item ClinicalT5 \cite{clinicalt5} is the T5 model \cite{t5} additionally fine-tuned on clinical text data including patient records and consultation reports. Harnessing its text-to-text transfer learning capability for multi-task modeling, ClinicalT5 can be applied to various natural language processing tasks in the healthcare domain, such as treatment classification, medical information extraction, and summarization of patient records.
\end{itemize}

For the process of encoding text concepts, we utilize the output from the BEiT model, which is specifically trained for the concept detection task. During this process, we apply a threshold of 0.5 to selectively retain predictions with a confidence score higher than 0.5, while discarding predictions with lower confidence scores. This discriminative process aids in capturing the semantic essence of the detected concepts, thereby facilitating their seamless integration into the multimodal fusion architecture for further processing and analysis.

\subsection{Experimental Settings}
Several experiments have been conducted to assess the efficacy of the proposed methodologies in addressing the ImageCLEFmedical Caption 2024 challenge. Specifically, each pre-trained vision model has been instantiated and evaluated, as detailed in Table 2, which offers a comprehensive overview of the pre-trained models employed in this study, encompassing their respective vision model designations, versions, and parameter counts for each fusion model. These experiments serve to elucidate both the potential and limitations inherent in each model with regard to the Image Captioning task, thereby facilitating the selection of the optimal approach for generating final predictions on the private test dataset of the competition.

\begin{itemize}
    \item 
\textbf{Concept Detection Task:}
For the concept detection subtask, the optimization criterion utilized during training is the AdamW optimizer \cite{kingma2015adam}. The models are trained for 5 epochs with a batch size of 30 and an initial learning rate of 5e-5. During training, the BCEWithLogitsLoss function, which combines a Sigmoid layer and BCELoss, is applied, and a threshold value ranging from 0.45 to 0.5 is predominantly used to process the model's output. To ensure meaningful comparison results, consistent hyperparameters are maintained across all experiments.

\item \textbf{Caption Prediction Task:}
During the training process for the caption prediction task, the CrossEntropyLoss criterion is applied with the ignore\_index parameter set to the pad token index of the tokenizer. This setup helps mitigate the influence of pad tokens on loss computation, ensuring more precise training outcomes. For optimization, the AdamW optimizer is utilized with a learning rate of 1e-4 and a weight decay rate of 0.01, chosen to balance training efficiency and model generalization \cite{kingma2015adam}.
To leverage the benefits of Mixed Precision Training \cite{micikevicius2018mixed}, the Gradient scaler is integrated into the training pipeline. This scaler adjusts the gradient scale, enhancing training efficiency and convergence speed of the models. Additionally, the LinearScheduleWithWarmup is employed to adjust the learning rate over time during training. This scheduling mechanism requires pre-defining the number of warmup steps and total training steps to optimize the learning rate schedule effectively.
During each training iteration, a batch size of 16 is utilized. Overall, these training configurations and optimizations contribute to the performance and stability of the training process, leading to superior model performance.
\end{itemize}

\subsection{Evaluation Methodology}

Our evaluation consists of two tasks: Concept Detection and Caption Prediction. Each task uses specific metrics to measure performance.
\begin{itemize}
    \item \textbf{Concept Detection Task:}
We assess the accuracy of concept identification using Accuracy, Precision, Recall, and F1 Score. These metrics measure overall correctness, positive prediction accuracy, relevant concept capture, and balanced precision and recall, respectively \cite{sokolova2009systematic}.

\item \textbf{Caption Prediction Task:}
We evaluate the quality and coherence of generated captions using BERT Score \cite{zhang2019bertscore}, BLEU (1-4) \cite{papineni2002bleu}, ROUGE Score \cite{lin2004rouge}, and METEOR Score \cite{banerjee2005meteor}. These metrics assess semantic similarity, fluency, relevance, coherence, informativeness, and lexical/syntactic aspects.
\end{itemize}

Using this diverse set of metrics, we ensure a comprehensive understanding of the model's performance and facilitate informed decision-making for further refinement.

\subsection{Experimental Results}

\begin{table}[h]
\caption{Comparative performance of the Concept Detection method on the validation set.}
\label{tab:concept_2}
\centering
\begin{tabular}{lcccc}
\hline
\textbf{Models}     & \textbf{Accuracy} & \textbf{Precision} & \textbf{Recall} & \textbf{F1} \\ \hline
Resnet-50  & 0.11412  & 0.89235   & 0.39643 & 0.51566 \\
BEiT-B     & 0.15554  & 0.93087   & 0.45961 & 0.57662 \\
Swin-V2    & \textbf{0.16366} & 0.94428   & \textbf{0.47114} & \textbf{0.58944} \\
DeiT-B     & 0.15674  & 0.93353   & 0.45849 & 0.57641 \\
ViT-B      & 0.15413  & 0.93477   & 0.45571 & 0.57439 \\
BiomedCLIP & 0.15975  & 0.94095   & 0.46453 & 0.58319 \\
BEIT-L     & 0.16145  & 0.93669   & 0.46700 & 0.58418 \\
Ensemble-2 & 0.16155  & 0.94501   & 0.46683 & 0.58581 \\
Ensemble-4 & 0.16135  & \textbf{0.94508}   & 0.46526 & 0.58460 \\ \hline
\end{tabular}
\end{table}

As detailed in Table~\ref{tab:concept_2}, the comparative evaluation of various concept detection models on the development validation set yields valuable insights into their performance across diverse evaluation metrics. Among these models, Swin-V2 emerges as the frontrunner, exhibiting the highest accuracy (0.16366), recall (0.47114), and F1 score (0.58944). This underscores Swin-V2's effectiveness in not only accurately identifying pertinent instances but also striking a harmonious balance between precision and recall, rendering it well-suited for concept detection endeavors.
Ensemble methodologies, which predictions from multiple models, demonstrate promising outcomes as well. Notably, the Ensemble-2 model showcases commendable precision (0.94501) and a noteworthy F1 score (0.58581), suggesting that leveraging diverse models can augment predictive efficacy, particularly in precision-oriented tasks. While the Ensemble-4 model marginally surpasses Ensemble-2 in precision (0.94508), it exhibits a slightly lower F1 score (0.58460), implying a subtle trade-off in recall when employing additional models.

BEiT-L and BiomedCLIP also manifest robust performance metrics. BEiT-L achieves an accuracy of 0.16145 and an F1 score of 0.58418, while BiomedCLIP demonstrates balanced performance with an accuracy of 0.15975 and an F1 score of 0.58319. These findings underscore the efficacy of these models in maintaining high precision and achieving a favorable balance with recall.

Other models such as BEiT-B, DeiT-B, and ViT-B exhibit commendable performance, albeit slightly trailing the top performers. For instance, BEiT-B records an accuracy of 0.15554 and an F1 score of 0.57662, indicating respectable yet not leading-edge performance. Similarly, DeiT-B and ViT-B attain comparable results, with DeiT-B registering an accuracy of 0.15674 and an F1 score of 0.57641, and ViT-B yielding an accuracy of 0.15413 and an F1 score of 0.57439.
Conversely, ResNet-50 demonstrates notably inferior performance across all metrics, with an accuracy of 0.11412 and an F1 score of 0.51566. This underscores its relatively limited efficacy in the concept detection task.

In summation, the Swin-V2 model emerges as the most dependable choice for concept detection owing to its superior accuracy, recall, and F1 score. Ensemble methodologies, particularly Ensemble-2, exhibit robust performance, underscoring the advantages of model amalgamation. BEiT-L and BiomedCLIP offer balanced performance, rendering them viable alternatives. Meanwhile, ResNet-50's diminished performance suggests its lesser suitability for this specific task, underscoring the strides made by newer architectural advancements.

\begin{table}[h]
\caption{A comparative analysis of various configurations on the validation set, with "Process" denoting post-processing of output captions to mitigate repetition, and "Concepts" representing features potentially derived from the Concept Detection subtask.}
\label{tab:performance_comparison}
\centering
\resizebox{\textwidth}{!}{
\begin{tabular}{llccccccc}
\hline
\textbf{Model} & \textbf{Configuration} & \textbf{BERTScore} & \textbf{BLEU1} & \textbf{BLEU2} & \textbf{BLEU3} & \textbf{BLEU4} & \textbf{ROUGE} & \textbf{METEOR} \\ \hline
BEiT+BioBart &  Concepts+No-Process & \textbf{0.60589} & 0.03293 & 0.01019 & 0.00337 & 0.00040 & 0.10721 & 0.05673 \\
BEiT+BioBart &  Concepts+Process    & \textbf{0.60589} & 0.03293 & 0.01019 & 0.00337 & 0.00040 & 0.10721 & 0.05673 \\
BEiT+Clinical-T5 & Concepts+No-Process   & 0.45752 & 0.07408 & 0.03008 & 0.01244 & 0.00476 & 0.09298 & 0.08909 \\
BEiT+Clinical-T5 & Concepts+Process      & 0.57597 & \textbf{0.08145} & \textbf{0.03319} & \textbf{0.01423} & \textbf{0.00519} & \textbf{0.13336} & 0.09817 \\
BEiT+Clinical-T5 & No-Concepts+No-Process & 0.46001 & 0.07501 & 0.03057 & 0.01077 & 0.00303 & 0.09711 & 0.09298 \\
BEiT+Clinical-T5 &  No-Concepts+Process   & 0.57487 & 0.08110 & 0.03231 & 0.01137 & 0.00310 & 0.13293 & \textbf{0.10086} \\ \hline
\end{tabular}
}
\end{table}

As detailed in Table~\ref{tab:performance_comparison}, the comparative analysis of various model configurations on the validation set reveals significant insights into the efficacy of incorporating concepts and post-processing techniques in caption generation tasks. The models evaluated include BEiT+BioBart and BEiT+Clinical-T5, with configurations either incorporating concepts derived from the Concept Detection subtask or excluding them, and applying post-processing to mitigate repetition in the output captions.
The results indicate that for the BEiT+BioBart model, the inclusion of concepts and the application of post-processing do not result in any variation in performance across all evaluated metrics, including BERTScore, BLEU (from 1 to 4), ROUGE, and METEOR. This suggests that for BEiT+BioBart, the post-processing step does not impact the model's ability to generate captions when concepts are included, maintaining consistent performance.

In contrast, the BEiT+Clinical-T5 model demonstrates a more nuanced response to the incorporation of concepts and post-processing. When concepts are included without post-processing, there is a slight decline in BERTScore compared to the configuration without concepts. However, BLEU, ROUGE, and METEOR scores show an improvement with the inclusion of concepts, highlighting the potential benefits of concept integration in enhancing the model's performance in these specific metrics. Notably, when post-processing is applied, the BEiT+Clinical-T5 model exhibits substantial improvements across all metrics, irrespective of the presence of concepts. This improvement underscores the critical role of post-processing in refining output quality, with the highest METEOR score observed in the configuration without concepts but with post-processing.
Comparing the two models, BEiT+Clinical-T5 generally outperforms BEiT+BioBart in BLEU, ROUGE, and METEOR scores. This superior performance is particularly evident when post-processing is applied, suggesting that BEiT+Clinical-T5 is more responsive to post-processing enhancements. However, BEiT+BioBart achieves a higher BERTScore when concepts are included, indicating a potential strength in semantic similarity measures.

In conclusion, the analysis underscores the importance of model selection, the strategic inclusion of concepts, and the application of post-processing in optimizing caption generation performance. BEiT+Clinical-T5 emerges as a more robust model with significant gains from post-processing, while BEiT+BioBart maintains consistent performance with concept inclusion. These findings provide valuable insights for future research and development in automated caption generation systems, emphasizing tailored approaches for different model architectures.

% \usepackage{booktabs}
\begin{table}[h]
\caption{Performance evaluation of different models on the validation set and private test set}
\label{tab:my-table-val-test-set}
\resizebox{0.7\textwidth}{!}{
\begin{tabular}{@{}lllcc@{}}
\toprule
\textbf{\#}         & \textbf{Models} & \textbf{Configuration}                         & \textbf{Validation set}         & \textbf{Test set}                    \\ \midrule
\textbf{Concept}    & BEiT-B           & \cellcolor[HTML]{FFFFFF}Threshold\_0.45        & 0.57662                         & \textbf{0.61079}                        \\
\textbf{Detection}  & BEiT-B           & \cellcolor[HTML]{FFFFFF}Threshold\_0.5         & -                               & 0.60904                                 \\
                    & Swin-V2          & \cellcolor[HTML]{FFFFFF}Threshold\_0.5         & 0.58944                         & 0.61998                                 \\ \midrule
\textbf{Caption}    & BEiT+Clinical-T5 & \cellcolor[HTML]{FFFFFF}No-Concepts+No-Process & \cellcolor[HTML]{FFFFFF}0.46001 & \cellcolor[HTML]{FFFFFF}0.4433          \\
\textbf{Prediction} & BEiT+Clinical-T5 & \cellcolor[HTML]{FFFFFF}Concepts+No-Process    & \cellcolor[HTML]{FFFFFF}0.45752 & 0.4453                                  \\
                    & BEiT+Clinical-T5 & \cellcolor[HTML]{FFFFFF}Concepts+Process       & \cellcolor[HTML]{FFFFFF}0.57597 & \cellcolor[HTML]{FFFFFF}0.558           \\
                    & BEiT+BioBart     & Concepts+Process                               & \cellcolor[HTML]{FFFFFF}0.60589 & \cellcolor[HTML]{FFFFFF}\textbf{0.5794} \\ \bottomrule
\end{tabular}
}
\end{table}

As detailed in Table~\ref{tab:my-table-val-test-set}, the performance evaluation of different models on the validation and private test sets provides a comprehensive understanding of their effectiveness across various configurations and datasets. For concept detection, three configurations were assessed: Concept BEiT-B with a threshold of 0.45, Detection BEiT-B with a threshold of 0.5, and Swin-V2 with a threshold of 0.5. The results reveal that the Swin-V2 model performs the best, achieving scores of 0.58944 on the validation set and 0.61998 on the private test set, suggesting superior capability in accurately detecting concepts compared to the BEiT-B models. The Concept BEiT-B model with a threshold of 0.45 also shows strong performance, though slightly lower than Swin-V2, indicating the threshold setting's impact on model efficacy.

For caption prediction, four configurations were evaluated: BEiT+Clinical-T5 without concepts and without post-processing, BEiT+Clinical-T5 with concepts and without post-processing, BEiT+Clinical-T5 with concepts and with post-processing, and BEiT+BioBart with concepts and with post-processing. The BEiT+Clinical-T5 model without concepts and post-processing scored 0.46001 on the validation set and 0.4433 on the private test set, while adding concepts slightly improved the private test set score to 0.4453. However, the most significant performance boost was observed when post-processing was applied to the BEiT+Clinical-T5 model with concepts, raising the scores to 0.57597 on the validation set and 0.558 on the private test set. This highlights the substantial role of post-processing in enhancing model performance.

Moreover, the BEiT+BioBart model with concepts and post-processing achieved the highest scores among all configurations, with 0.60589 on the validation set and 0.5794 on the private test set. This underscores the effectiveness of combining concepts with post-processing in the BioBart architecture, suggesting that such integration can significantly improve caption generation quality.
Overall, the analysis emphasizes the critical influence of model configuration, the integration of concepts, and the application of post-processing on the performance outcomes. The superior performance of the Swin-V2 model for concept detection and the BEiT+BioBart model for caption prediction indicates that different models may excel in specific sub-tasks, advocating for a nuanced approach in model selection and optimization based on the task requirements and dataset characteristics.

\subsection{Error analysis}

\begin{figure}[h]
    \centering
    \includegraphics[width=0.75\linewidth]{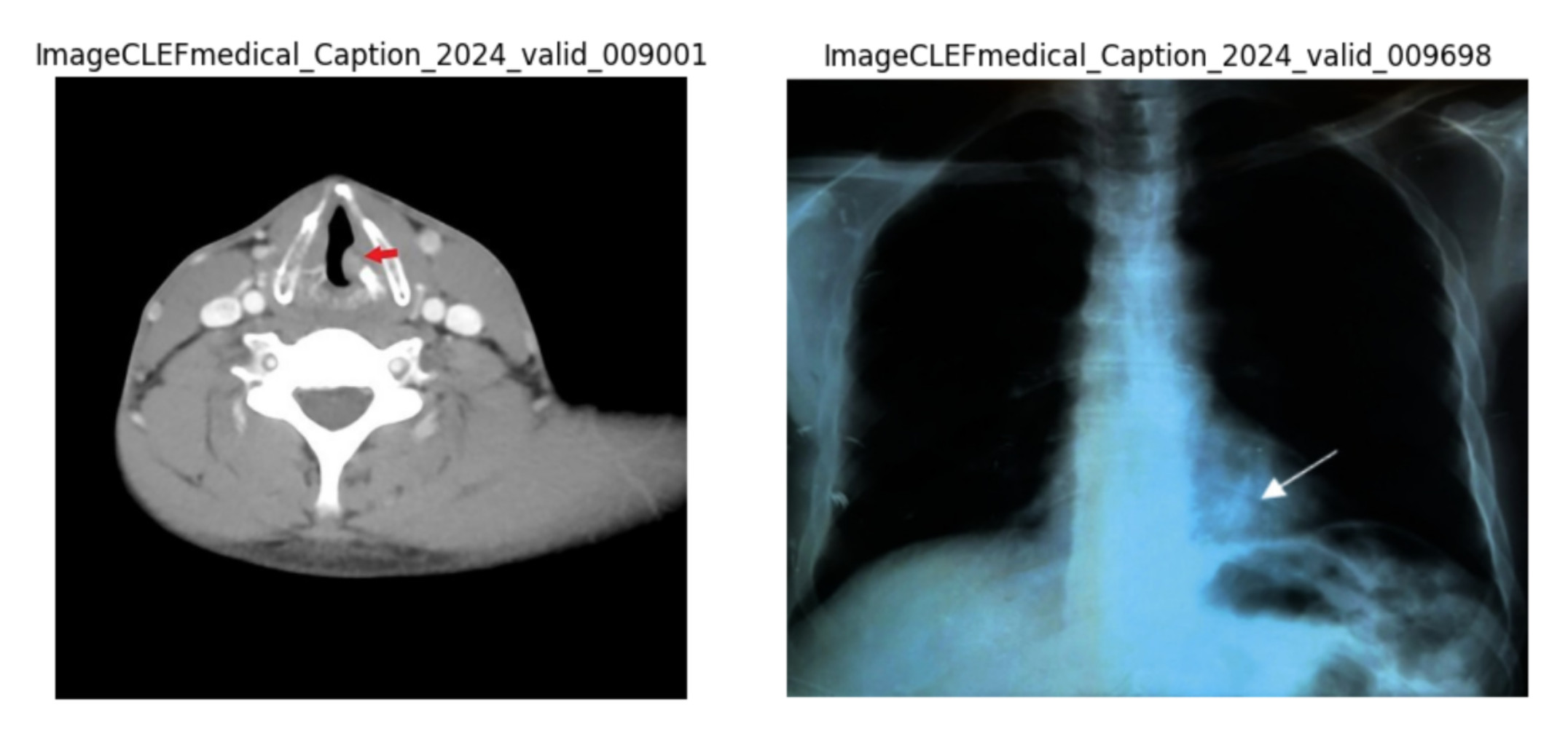}
    \caption{Example images of caption prediction. The images are arranged in sequential order in Table~\ref{tab:my-table-err}. \\ a) \texttt{ImageCLEFmedical\_Caption\_2024\_valid\_009001} is an example of Table~\ref{tab:my-table-err}, from the dataset provided under Hijazi2022. \\b) \texttt{ImageCLEFmedical\_Caption\_2024\_valid\_009698} is an example of Table~\ref{tab:my-table-err}, from the dataset provided under Khaled2022.}
    \label{fig:enter-label}
\end{figure}

% \usepackage{booktabs}
\begin{table}[h]
\caption{Example outputs of caption prediction from different models.}
\label{tab:my-table-err}
\resizebox{0.8\textwidth}{!}{
\begin{tabular}{@{}lccc@{}}
\toprule

\textbf{Ground Truth}                                                                                                                  & \textbf{BEiT + Clinical-T5}                                                                   & \textbf{BEiT + BioBART}                                                              & \textbf{\begin{tabular}[c]{@{}c@{}}Predicted \\ Concepts\end{tabular}}   \\ \midrule
\begin{tabular}[c]{@{}l@{}}Axial contrasted CT image of \\ larynx, showing left sided glottic\\ versus supraglottic mass.\end{tabular} & \begin{tabular}[c]{@{}c@{}}CT scan showing mass\\ lesion (arrow)\end{tabular}                 & \begin{tabular}[c]{@{}c@{}}CT scan showing \\ left renal mass\end{tabular}           & \begin{tabular}[c]{@{}c@{}}Magnetic \\ Resonance \\ Imaging\end{tabular} \\ \midrule
\begin{tabular}[c]{@{}l@{}}Chest X-ray face (solitary \\ pulmonary nodule of the \\ heart-phrenic angle).\end{tabular}                 & \begin{tabular}[c]{@{}c@{}}Chest X-ray showing \\ opacification \\ (arrow) chest\end{tabular} & \begin{tabular}[c]{@{}c@{}}Chest X-ray showing \\ bilateral infiltrates\end{tabular} & \begin{tabular}[c]{@{}c@{}}X-Ray Computed \\ Tomography\end{tabular}     \\ \bottomrule
\end{tabular}
}
\end{table}

As detailed in Table~\ref{tab:my-table-err}, when employing the BEiT model in conjunction with ClinicalT5 for medical image analysis, several notable errors have been observed across various dimensions. These errors include incorrect identification of regions or image types, omissions in providing specific details, and inaccuracies in context, thereby impacting the overall reliability of the model's results. 
The model occasionally encounters difficulties in accurately identifying regions of interest within the images. For instance, it might misinterpret an anteroposterior X-ray of the pelvis as indicating bilateral tibial fractures. Similarly, it might incorrectly classify a cross-sectional, contrast-enhanced CT scan of the larynx as a left renal tumor.

Omissions in providing specific details have become evident in the model's predictions. The model often fails to provide the complex details necessary for comprehensive clinical interpretation. For example, it may overlook critical features such as the eccentric position of a metallic head in an X-ray or the presence of stratified bile in a CT scan.
Moreover, contextual inaccuracies are common, leading to misleading or entirely incorrect descriptions. The model sometimes struggles to grasp the broader context of medical images, resulting in descriptions that do not align appropriately with the actual content of the images.
Similarly, when utilizing the BEiT model in combination with BioBART, analogous errors have been observed across various aspects. These include incorrect identification of regions or image types, omissions in providing specific details, and contextual inaccuracies. Comparing BEiT with ClinicalT5 and BEiT with BioBART, although both models exhibit similar error patterns, there are minor differences in their performance. BEiT combined with ClinicalT5 demonstrates slightly better performance in certain aspects, such as providing more accurate descriptions and better contextual understanding. Conversely, BEiT combined with BioBART shows a slight advantage in specific scenarios, particularly in identifying anatomical structures or image types. However, both models have room for improvement, highlighting ongoing challenges in developing robust and reliable automated methods for medical image analysis.
In both models, conceptual errors frequently occur, indicating a mismatch between the predicted concept and the actual content of the medical images. These errors underscore the challenges in accurately interpreting and classifying medical images based on their content.

To enhance the accuracy of medical image analysis models, a range of strategies must be employed to improve data quality, model architecture, and training processes. Firstly, the use of high-quality, well-annotated datasets is crucial. Combining this with data augmentation techniques such as rotation, zooming, flipping, and color adjustment can help increase the size and diversity of the training dataset, thereby enhancing the model's generalization capabilities. In terms of model architecture, employing models pre-trained on domain-specific datasets or state-of-the-art (SOTA) models that achieve superior results is essential. Furthermore, incorporating additional feature extraction from image data, such as bounding-boxes, segmentation, or advanced features, can help the model better understand the structure and context of the images. Finally, regularly testing and re-evaluating the model using diverse datasets will help in early detection of errors and timely adjustment of the model, ensuring the reliability and accuracy of medical image analysis results.

\section{Conclusion and Future Works}
In this study, we introduced an enhanced approach to medical caption generation by integrating concept detection into attention mechanisms. Our method significantly improved performance metrics, with the Swin-V2 model achieving an F1 score of 0.58944 on the validation set and 0.61998 on the private test set, earning 3rd place in concept detection. For caption prediction, the BEiT+BioBart model, augmented with concept integration and post-processing, achieved a BERTScore of 0.60589 on the validation set and 0.5794 on the private test set, securing 9th place. These results underscore the effectiveness of concept-aware systems in generating precise and contextually relevant medical descriptions.

Future work will focus on enhancing model performance through several avenues unrelated to data expansion. First, optimizing model architectures and training protocols can further improve accuracy and efficiency. Second, incorporating more advanced attention mechanisms and fine-tuning hyperparameters may yield better contextual understanding and caption quality. Third, integrating explainability techniques will ensure that model predictions are interpretable and trustworthy for healthcare professionals. Additionally, exploring transfer learning and domain adaptation techniques could enhance model performance across various medical imaging modalities. 
Furthermore, leveraging large language models (LLMs) such as GPT-3 and BioGPT for their potential to generate coherent and diagnostically valuable medical captions will be explored \cite{Brown} \cite{Lou}. Finally, we plan to develop robust post-processing algorithms to further refine generated captions, ensuring they meet clinical standards. These efforts aim to advance the capabilities of medical image analysis and automated reporting systems, contributing to more sophisticated and reliable tools for the healthcare industry.


\section*{Acknowledgment}\label{secA1}
This research was supported by The VNUHCM-University of Information Technology's Scientific Research Support Fund.
%%
%% Define the bibliography file to be used
\bibliography{sample-ceur}

\begin{thebibliography}{39}
\expandafter\ifx\csname natexlab\endcsname\relax\def\natexlab#1{#1}\fi
\providecommand{\url}[1]{\texttt{#1}}
\providecommand{\href}[2]{#2}
\providecommand{\path}[1]{#1}
\providecommand{\DOIprefix}{doi:}
\providecommand{\ArXivprefix}{arXiv:}
\providecommand{\URLprefix}{URL: }
\providecommand{\Pubmedprefix}{pmid:}
\providecommand{\doi}[1]{\href{http://dx.doi.org/#1}{\path{#1}}}
\providecommand{\Pubmed}[1]{\href{pmid:#1}{\path{#1}}}
\providecommand{\bibinfo}[2]{#2}
\ifx\xfnm\relax \def\xfnm[#1]{\unskip,\space#1}\fi
%Type = Article
\bibitem[{Litjens et~al.(2017)Litjens, Kooi, Bejnordi, Setio, Ciompi, Ghafoorian, {van der Laak}, {van Ginneken}, and Sánchez}]{intro0}
\bibinfo{author}{G.~Litjens}, \bibinfo{author}{T.~Kooi}, \bibinfo{author}{B.~E. Bejnordi}, \bibinfo{author}{A.~A.~A. Setio}, \bibinfo{author}{F.~Ciompi}, \bibinfo{author}{M.~Ghafoorian}, \bibinfo{author}{J.~A. {van der Laak}}, \bibinfo{author}{B.~{van Ginneken}}, \bibinfo{author}{C.~I. Sánchez},
\newblock \bibinfo{title}{A survey on deep learning in medical image analysis},
\newblock \bibinfo{journal}{Medical Image Analysis} \bibinfo{volume}{42} (\bibinfo{year}{2017}) \bibinfo{pages}{60--88}. \URLprefix \url{https://www.sciencedirect.com/science/article/pii/S1361841517301135}. \DOIprefix\doi{https://doi.org/10.1016/j.media.2017.07.005}.
%Type = Article
\bibitem[{Esteva et~al.(2017)Esteva, Kuprel, Novoa, Ko, Swetter, Blau, and Thrun}]{intro0.1}
\bibinfo{author}{A.~Esteva}, \bibinfo{author}{B.~Kuprel}, \bibinfo{author}{R.~A. Novoa}, \bibinfo{author}{J.~Ko}, \bibinfo{author}{S.~M. Swetter}, \bibinfo{author}{H.~M. Blau}, \bibinfo{author}{S.~Thrun},
\newblock \bibinfo{title}{Dermatologist-level classification of skin cancer with deep neural networks},
\newblock \bibinfo{journal}{nature} \bibinfo{volume}{542} (\bibinfo{year}{2017}) \bibinfo{pages}{115--118}.
%Type = Inproceedings
\bibitem[{Jing et~al.(2018)Jing, Xie, and Xing}]{intro1}
\bibinfo{author}{B.~Jing}, \bibinfo{author}{P.~Xie}, \bibinfo{author}{E.~Xing},
\newblock \bibinfo{title}{On the automatic generation of medical imaging reports},
\newblock in: \bibinfo{editor}{I.~Gurevych}, \bibinfo{editor}{Y.~Miyao} (Eds.), \bibinfo{booktitle}{Proceedings of the 56th Annual Meeting of the Association for Computational Linguistics (Volume 1: Long Papers)}, \bibinfo{publisher}{Association for Computational Linguistics}, \bibinfo{address}{Melbourne, Australia}, \bibinfo{year}{2018}, pp. \bibinfo{pages}{2577--2586}. \URLprefix \url{https://aclanthology.org/P18-1240}. \DOIprefix\doi{10.18653/v1/P18-1240}.
%Type = Inproceedings
\bibitem[{Li et~al.(2019)Li, Liang, Hu, and Xing}]{intro2}
\bibinfo{author}{C.~Y. Li}, \bibinfo{author}{X.~Liang}, \bibinfo{author}{Z.~Hu}, \bibinfo{author}{E.~P. Xing},
\newblock \bibinfo{title}{Knowledge-driven encode, retrieve, paraphrase for medical image report generation},
\newblock in: \bibinfo{booktitle}{Proceedings of the Thirty-Third AAAI Conference on Artificial Intelligence and Thirty-First Innovative Applications of Artificial Intelligence Conference and Ninth AAAI Symposium on Educational Advances in Artificial Intelligence}, AAAI'19/IAAI'19/EAAI'19, \bibinfo{publisher}{AAAI Press}, \bibinfo{year}{2019}. \URLprefix \url{https://doi.org/10.1609/aaai.v33i01.33016666}. \DOIprefix\doi{10.1609/aaai.v33i01.33016666}.
%Type = Inproceedings
\bibitem[{Shin et~al.(2016)Shin, Roberts, Lu, Demner-Fushman, Yao, and Summers}]{intro3}
\bibinfo{author}{H.-C. Shin}, \bibinfo{author}{K.~Roberts}, \bibinfo{author}{L.~Lu}, \bibinfo{author}{D.~Demner-Fushman}, \bibinfo{author}{J.~Yao}, \bibinfo{author}{R.~M. Summers},
\newblock \bibinfo{title}{Learning to read chest x-rays: Recurrent neural cascade model for automated image annotation},
\newblock in: \bibinfo{booktitle}{2016 IEEE Conference on Computer Vision and Pattern Recognition (CVPR)}, \bibinfo{year}{2016}, pp. \bibinfo{pages}{2497--2506}. \DOIprefix\doi{10.1109/CVPR.2016.274}.
%Type = Inproceedings
\bibitem[{Ionescu et~al.(2024)Ionescu, M"uller, Dr\u{a}gulinescu, R"uckert, {Ben Abacha}, {Garc\'{\i}a Seco de Herrera}, Bloch, Br"ungel, Idrissi{-}Yaghir, Sch"afer, Schmidt, Pakull, Damm, Bracke, Friedrich, Andrei, Prokopchuk, Karpenka, Radzhabov, Kovalev, Macaire, Schwab, Lecouteux, Esperan\c{c}a{-}Rodier, Yim, Fu, Sun, Yetisgen, Xia, Hicks, Riegler, Thambawita, Stor\r{a}s, Halvorsen, Heinrich, Kiesel, Potthast, and Stein}]{ImageCLEF2024}
\bibinfo{author}{B.~Ionescu}, \bibinfo{author}{H.~M"uller}, \bibinfo{author}{A.~Dr\u{a}gulinescu}, \bibinfo{author}{J.~R"uckert}, \bibinfo{author}{A.~{Ben Abacha}}, \bibinfo{author}{A.~{Garc\'{\i}a Seco de Herrera}}, \bibinfo{author}{L.~Bloch}, \bibinfo{author}{R.~Br"ungel}, \bibinfo{author}{A.~Idrissi{-}Yaghir}, \bibinfo{author}{H.~Sch"afer}, \bibinfo{author}{C.~S. Schmidt}, \bibinfo{author}{T.~M.~G. Pakull}, \bibinfo{author}{H.~Damm}, \bibinfo{author}{B.~Bracke}, \bibinfo{author}{C.~M. Friedrich}, \bibinfo{author}{A.~Andrei}, \bibinfo{author}{Y.~Prokopchuk}, \bibinfo{author}{D.~Karpenka}, \bibinfo{author}{A.~Radzhabov}, \bibinfo{author}{V.~Kovalev}, \bibinfo{author}{C.~Macaire}, \bibinfo{author}{D.~Schwab}, \bibinfo{author}{B.~Lecouteux}, \bibinfo{author}{E.~Esperan\c{c}a{-}Rodier}, \bibinfo{author}{W.~Yim}, \bibinfo{author}{Y.~Fu}, \bibinfo{author}{Z.~Sun}, \bibinfo{author}{M.~Yetisgen}, \bibinfo{author}{F.~Xia}, \bibinfo{author}{S.~A. Hicks}, \bibinfo{author}{M.~A. Riegler},
  \bibinfo{author}{V.~Thambawita}, \bibinfo{author}{A.~Stor\r{a}s}, \bibinfo{author}{P.~Halvorsen}, \bibinfo{author}{M.~Heinrich}, \bibinfo{author}{J.~Kiesel}, \bibinfo{author}{M.~Potthast}, \bibinfo{author}{B.~Stein},
\newblock \bibinfo{title}{{Overview of ImageCLEF 2024}: Multimedia retrieval in medical applications},
\newblock in: \bibinfo{booktitle}{Experimental IR Meets Multilinguality, Multimodality, and Interaction}, Proceedings of the 15th International Conference of the CLEF Association (CLEF 2024), \bibinfo{publisher}{Springer Lecture Notes in Computer Science LNCS}, \bibinfo{address}{Grenoble, France}, \bibinfo{year}{2024}.
%Type = Inproceedings
\bibitem[{R"uckert et~al.(2024)R"uckert, Ben~Abacha, G.~Seco~de Herrera, Bloch, Br"ungel, Idrissi-Yaghir, Sch"afer, Bracke, Damm, Pakull, Schmidt, M"uller, and Friedrich}]{ImageCLEFmedicalCaptionOverview2024}
\bibinfo{author}{J.~R"uckert}, \bibinfo{author}{A.~Ben~Abacha}, \bibinfo{author}{A.~G.~Seco~de Herrera}, \bibinfo{author}{L.~Bloch}, \bibinfo{author}{R.~Br"ungel}, \bibinfo{author}{A.~Idrissi-Yaghir}, \bibinfo{author}{H.~Sch"afer}, \bibinfo{author}{B.~Bracke}, \bibinfo{author}{H.~Damm}, \bibinfo{author}{T.~M.~G. Pakull}, \bibinfo{author}{C.~S. Schmidt}, \bibinfo{author}{H.~M"uller}, \bibinfo{author}{C.~M. Friedrich},
\newblock \bibinfo{title}{Overview of {ImageCLEFmedical} 2024 -- {Caption Prediction and Concept Detection}},
\newblock in: \bibinfo{booktitle}{CLEF2024 Working Notes}, {CEUR} Workshop Proceedings, \bibinfo{publisher}{CEUR-WS.org}, \bibinfo{address}{Grenoble, France}, \bibinfo{year}{2024}.
%Type = Inproceedings
\bibitem[{Wang et~al.(2017)Wang, Peng, Lu, Lu, Bagheri, and Summers}]{re1}
\bibinfo{author}{X.~Wang}, \bibinfo{author}{Y.~Peng}, \bibinfo{author}{L.~Lu}, \bibinfo{author}{Z.~Lu}, \bibinfo{author}{M.~Bagheri}, \bibinfo{author}{R.~M. Summers},
\newblock \bibinfo{title}{Chestx-ray8: Hospital-scale chest x-ray database and benchmarks on weakly-supervised classification and localization of common thorax diseases},
\newblock in: \bibinfo{booktitle}{2017 IEEE Conference on Computer Vision and Pattern Recognition (CVPR)}, \bibinfo{year}{2017}, pp. \bibinfo{pages}{3462--3471}. \DOIprefix\doi{10.1109/CVPR.2017.369}.
%Type = Article
\bibitem[{Goldberger et~al.(2000)Goldberger, Amaral, Glass, Hausdorff, Ivanov, Mark, Mietus, Moody, Peng, and Stanley}]{re2}
\bibinfo{author}{A.~L. Goldberger}, \bibinfo{author}{L.~A. Amaral}, \bibinfo{author}{L.~Glass}, \bibinfo{author}{J.~M. Hausdorff}, \bibinfo{author}{P.~C. Ivanov}, \bibinfo{author}{R.~G. Mark}, \bibinfo{author}{J.~E. Mietus}, \bibinfo{author}{G.~B. Moody}, \bibinfo{author}{C.-K. Peng}, \bibinfo{author}{H.~E. Stanley},
\newblock \bibinfo{title}{Physiobank, physiotoolkit, and physionet: components of a new research resource for complex physiologic signals},
\newblock \bibinfo{journal}{Circulation} \bibinfo{volume}{101} (\bibinfo{year}{2000}) \bibinfo{pages}{e215--e220}.
%Type = Inproceedings
\bibitem[{He et~al.(2016)He, Zhang, Ren, and Sun}]{re21}
\bibinfo{author}{K.~He}, \bibinfo{author}{X.~Zhang}, \bibinfo{author}{S.~Ren}, \bibinfo{author}{J.~Sun},
\newblock \bibinfo{title}{Deep residual learning for image recognition},
\newblock in: \bibinfo{booktitle}{Proceedings of the IEEE conference on computer vision and pattern recognition}, \bibinfo{year}{2016}, pp. \bibinfo{pages}{770--778}.
%Type = Inproceedings
\bibitem[{Tan and Le(2019)}]{re22}
\bibinfo{author}{M.~Tan}, \bibinfo{author}{Q.~Le},
\newblock \bibinfo{title}{Efficientnet: Rethinking model scaling for convolutional neural networks},
\newblock in: \bibinfo{booktitle}{International conference on machine learning}, \bibinfo{organization}{PMLR}, \bibinfo{year}{2019}, pp. \bibinfo{pages}{6105--6114}.
%Type = Article
\bibitem[{Dosovitskiy et~al.(2020)Dosovitskiy, Beyer, Kolesnikov, Weissenborn, Zhai, Unterthiner, Dehghani, Minderer, Heigold, Gelly et~al.}]{re23}
\bibinfo{author}{A.~Dosovitskiy}, \bibinfo{author}{L.~Beyer}, \bibinfo{author}{A.~Kolesnikov}, \bibinfo{author}{D.~Weissenborn}, \bibinfo{author}{X.~Zhai}, \bibinfo{author}{T.~Unterthiner}, \bibinfo{author}{M.~Dehghani}, \bibinfo{author}{M.~Minderer}, \bibinfo{author}{G.~Heigold}, \bibinfo{author}{S.~Gelly}, et~al.,
\newblock \bibinfo{title}{An image is worth 16x16 words: Transformers for image recognition at scale},
\newblock \bibinfo{journal}{arXiv preprint arXiv:2010.11929}  (\bibinfo{year}{2020}).
%Type = Article
\bibitem[{Bao et~al.(2021)Bao, Dong, Piao, and Wei}]{re24}
\bibinfo{author}{H.~Bao}, \bibinfo{author}{L.~Dong}, \bibinfo{author}{S.~Piao}, \bibinfo{author}{F.~Wei},
\newblock \bibinfo{title}{Beit: Bert pre-training of image transformers},
\newblock \bibinfo{journal}{arXiv preprint arXiv:2106.08254}  (\bibinfo{year}{2021}).
%Type = Inproceedings
\bibitem[{Liu et~al.(2021)Liu, Lin, Cao, Hu, Wei, Zhang, Lin, and Guo}]{re25}
\bibinfo{author}{Z.~Liu}, \bibinfo{author}{Y.~Lin}, \bibinfo{author}{Y.~Cao}, \bibinfo{author}{H.~Hu}, \bibinfo{author}{Y.~Wei}, \bibinfo{author}{Z.~Zhang}, \bibinfo{author}{S.~Lin}, \bibinfo{author}{B.~Guo},
\newblock \bibinfo{title}{Swin transformer: Hierarchical vision transformer using shifted windows},
\newblock in: \bibinfo{booktitle}{Proceedings of the IEEE/CVF international conference on computer vision}, \bibinfo{year}{2021}, pp. \bibinfo{pages}{10012--10022}.
%Type = Article
\bibitem[{Hochreiter and Schmidhuber(1997)}]{lstm}
\bibinfo{author}{S.~Hochreiter}, \bibinfo{author}{J.~Schmidhuber},
\newblock \bibinfo{title}{Long short-term memory},
\newblock \bibinfo{journal}{Neural Comput.} \bibinfo{volume}{9} (\bibinfo{year}{1997}) \bibinfo{pages}{1735–1780}. \URLprefix \url{https://doi.org/10.1162/neco.1997.9.8.1735}. \DOIprefix\doi{10.1162/neco.1997.9.8.1735}.
%Type = Article
\bibitem[{Lee et~al.(2020)Lee, Yoon, Kim, Kim, Kim, So, and Kang}]{re32}
\bibinfo{author}{J.~Lee}, \bibinfo{author}{W.~Yoon}, \bibinfo{author}{S.~Kim}, \bibinfo{author}{D.~Kim}, \bibinfo{author}{S.~Kim}, \bibinfo{author}{C.~H. So}, \bibinfo{author}{J.~Kang},
\newblock \bibinfo{title}{Biobert: a pre-trained biomedical language representation model for biomedical text mining},
\newblock \bibinfo{journal}{Bioinformatics} \bibinfo{volume}{36} (\bibinfo{year}{2020}) \bibinfo{pages}{1234--1240}.
%Type = Article
\bibitem[{Luo et~al.(2022)Luo, Sun, Xia, Qin, Zhang, Poon, and Liu}]{re34}
\bibinfo{author}{R.~Luo}, \bibinfo{author}{L.~Sun}, \bibinfo{author}{Y.~Xia}, \bibinfo{author}{T.~Qin}, \bibinfo{author}{S.~Zhang}, \bibinfo{author}{H.~Poon}, \bibinfo{author}{T.-Y. Liu},
\newblock \bibinfo{title}{Biogpt: generative pre-trained transformer for biomedical text generation and mining},
\newblock \bibinfo{journal}{Briefings in bioinformatics} \bibinfo{volume}{23} (\bibinfo{year}{2022}) \bibinfo{pages}{bbac409}.
%Type = Misc
\bibitem[{Rückert et~al.(2024)Rückert, Bloch, Brüngel, Idrissi-Yaghir, Schäfer, Schmidt, Koitka, Pelka, Abacha, de~Herrera, Müller, Horn, Nensa, and Friedrich}]{rocov2}
\bibinfo{author}{J.~Rückert}, \bibinfo{author}{L.~Bloch}, \bibinfo{author}{R.~Brüngel}, \bibinfo{author}{A.~Idrissi-Yaghir}, \bibinfo{author}{H.~Schäfer}, \bibinfo{author}{C.~S. Schmidt}, \bibinfo{author}{S.~Koitka}, \bibinfo{author}{O.~Pelka}, \bibinfo{author}{A.~B. Abacha}, \bibinfo{author}{A.~G.~S. de~Herrera}, \bibinfo{author}{H.~Müller}, \bibinfo{author}{P.~A. Horn}, \bibinfo{author}{F.~Nensa}, \bibinfo{author}{C.~M. Friedrich}, \bibinfo{title}{{ROCOv2}: {Radiology Objects in COntext} version 2, an updated multimodal image dataset}, \bibinfo{year}{2024}. \URLprefix \url{https://arxiv.org/abs/2405.10004v1}. \href{http://arxiv.org/abs/2405.10004}{{\tt arXiv:2405.10004}}.
%Type = Inproceedings
\bibitem[{Pelka et~al.(2018)Pelka, Koitka, R{\"u}ckert, Nensa, and Friedrich}]{roco}
\bibinfo{author}{O.~Pelka}, \bibinfo{author}{S.~Koitka}, \bibinfo{author}{J.~R{\"u}ckert}, \bibinfo{author}{F.~Nensa}, \bibinfo{author}{C.~M. Friedrich},
\newblock \bibinfo{title}{Radiology objects in context (roco): a multimodal image dataset},
\newblock in: \bibinfo{booktitle}{Intravascular Imaging and Computer Assisted Stenting and Large-Scale Annotation of Biomedical Data and Expert Label Synthesis: 7th Joint International Workshop, CVII-STENT 2018 and Third International Workshop, LABELS 2018, Held in Conjunction with MICCAI 2018, Granada, Spain, September 16, 2018, Proceedings 3}, \bibinfo{organization}{Springer}, \bibinfo{year}{2018}, pp. \bibinfo{pages}{180--189}.
%Type = Misc
\bibitem[{{National Library of Medicine}(2003)}]{pmc}
\bibinfo{author}{{National Library of Medicine}}, \bibinfo{title}{Pmc open access subset}, \bibinfo{howpublished}{\url{https://www.ncbi.nlm.nih.gov/pmc/tools/openftlist/}}, \bibinfo{year}{2003}. \bibinfo{note}{[cited 2024 May 30]}.
%Type = Inproceedings
\bibitem[{He et~al.(2016)He, Zhang, Ren, and Sun}]{he2016deep}
\bibinfo{author}{K.~He}, \bibinfo{author}{X.~Zhang}, \bibinfo{author}{S.~Ren}, \bibinfo{author}{J.~Sun},
\newblock \bibinfo{title}{Deep residual learning for image recognition},
\newblock in: \bibinfo{booktitle}{Proceedings of the IEEE conference on computer vision and pattern recognition}, \bibinfo{year}{2016}, pp. \bibinfo{pages}{770--778}.
%Type = Misc
\bibitem[{Dosovitskiy et~al.(2021)Dosovitskiy, Beyer, Kolesnikov, Weissenborn, Zhai, Unterthiner, Dehghani, Minderer, Heigold, Gelly, Uszkoreit, and Houlsby}]{dosovitskiy2021image}
\bibinfo{author}{A.~Dosovitskiy}, \bibinfo{author}{L.~Beyer}, \bibinfo{author}{A.~Kolesnikov}, \bibinfo{author}{D.~Weissenborn}, \bibinfo{author}{X.~Zhai}, \bibinfo{author}{T.~Unterthiner}, \bibinfo{author}{M.~Dehghani}, \bibinfo{author}{M.~Minderer}, \bibinfo{author}{G.~Heigold}, \bibinfo{author}{S.~Gelly}, \bibinfo{author}{J.~Uszkoreit}, \bibinfo{author}{N.~Houlsby}, \bibinfo{title}{An image is worth 16x16 words: Transformers for image recognition at scale}, \bibinfo{year}{2021}. \href{http://arxiv.org/abs/2010.11929}{{\tt arXiv:2010.11929}}.
%Type = Misc
\bibitem[{Touvron et~al.(2021)Touvron, Cord, Douze, Massa, Sablayrolles, and Jégou}]{touvron2021training}
\bibinfo{author}{H.~Touvron}, \bibinfo{author}{M.~Cord}, \bibinfo{author}{M.~Douze}, \bibinfo{author}{F.~Massa}, \bibinfo{author}{A.~Sablayrolles}, \bibinfo{author}{H.~Jégou}, \bibinfo{title}{Training data-efficient image transformers \& distillation through attention}, \bibinfo{year}{2021}. \href{http://arxiv.org/abs/2012.12877}{{\tt arXiv:2012.12877}}.
%Type = Misc
\bibitem[{Liu et~al.(2022)Liu, Hu, Lin, Yao, Xie, Wei, Ning, Cao, Zhang, Dong, Wei, and Guo}]{swinv2}
\bibinfo{author}{Z.~Liu}, \bibinfo{author}{H.~Hu}, \bibinfo{author}{Y.~Lin}, \bibinfo{author}{Z.~Yao}, \bibinfo{author}{Z.~Xie}, \bibinfo{author}{Y.~Wei}, \bibinfo{author}{J.~Ning}, \bibinfo{author}{Y.~Cao}, \bibinfo{author}{Z.~Zhang}, \bibinfo{author}{L.~Dong}, \bibinfo{author}{F.~Wei}, \bibinfo{author}{B.~Guo}, \bibinfo{title}{Swin transformer v2: Scaling up capacity and resolution}, \bibinfo{year}{2022}. \href{http://arxiv.org/abs/2111.09883}{{\tt arXiv:2111.09883}}.
%Type = Article
\bibitem[{Bao et~al.(2021)Bao, Dong, and Wei}]{beit}
\bibinfo{author}{H.~Bao}, \bibinfo{author}{L.~Dong}, \bibinfo{author}{F.~Wei},
\newblock \bibinfo{title}{Beit: {BERT} pre-training of image transformers},
\newblock \bibinfo{journal}{CoRR} \bibinfo{volume}{abs/2106.08254} (\bibinfo{year}{2021}). \URLprefix \url{https://arxiv.org/abs/2106.08254}. \href{http://arxiv.org/abs/2106.08254}{{\tt arXiv:2106.08254}}.
%Type = Misc
\bibitem[{Zhang et~al.(2024)Zhang, Xu, Usuyama, Xu, Bagga, Tinn, Preston, Rao, Wei, Valluri, Wong, Tupini, Wang, Mazzola, Shukla, Liden, Gao, Lungren, Naumann, Wang, and Poon}]{biomedclip}
\bibinfo{author}{S.~Zhang}, \bibinfo{author}{Y.~Xu}, \bibinfo{author}{N.~Usuyama}, \bibinfo{author}{H.~Xu}, \bibinfo{author}{J.~Bagga}, \bibinfo{author}{R.~Tinn}, \bibinfo{author}{S.~Preston}, \bibinfo{author}{R.~Rao}, \bibinfo{author}{M.~Wei}, \bibinfo{author}{N.~Valluri}, \bibinfo{author}{C.~Wong}, \bibinfo{author}{A.~Tupini}, \bibinfo{author}{Y.~Wang}, \bibinfo{author}{M.~Mazzola}, \bibinfo{author}{S.~Shukla}, \bibinfo{author}{L.~Liden}, \bibinfo{author}{J.~Gao}, \bibinfo{author}{M.~P. Lungren}, \bibinfo{author}{T.~Naumann}, \bibinfo{author}{S.~Wang}, \bibinfo{author}{H.~Poon}, \bibinfo{title}{Biomedclip: a multimodal biomedical foundation model pretrained from fifteen million scientific image-text pairs}, \bibinfo{year}{2024}. \href{http://arxiv.org/abs/2303.00915}{{\tt arXiv:2303.00915}}.
%Type = Misc
\bibitem[{Yuan et~al.(2022)Yuan, Yuan, Gan, Zhang, Xie, and Yu}]{BioBART}
\bibinfo{author}{H.~Yuan}, \bibinfo{author}{Z.~Yuan}, \bibinfo{author}{R.~Gan}, \bibinfo{author}{J.~Zhang}, \bibinfo{author}{Y.~Xie}, \bibinfo{author}{S.~Yu}, \bibinfo{title}{Biobart: Pretraining and evaluation of a biomedical generative language model}, \bibinfo{year}{2022}. \href{http://arxiv.org/abs/2204.03905}{{\tt arXiv:2204.03905}}.
%Type = Inproceedings
\bibitem[{Lewis et~al.(2020)Lewis, Liu, Goyal, Ghazvininejad, Mohamed, Levy, Stoyanov, and Zettlemoyer}]{bart}
\bibinfo{author}{M.~Lewis}, \bibinfo{author}{Y.~Liu}, \bibinfo{author}{N.~Goyal}, \bibinfo{author}{M.~Ghazvininejad}, \bibinfo{author}{A.~Mohamed}, \bibinfo{author}{O.~Levy}, \bibinfo{author}{V.~Stoyanov}, \bibinfo{author}{L.~Zettlemoyer},
\newblock \bibinfo{title}{{BART}: Denoising sequence-to-sequence pre-training for natural language generation, translation, and comprehension},
\newblock in: \bibinfo{editor}{D.~Jurafsky}, \bibinfo{editor}{J.~Chai}, \bibinfo{editor}{N.~Schluter}, \bibinfo{editor}{J.~Tetreault} (Eds.), \bibinfo{booktitle}{Proceedings of the 58th Annual Meeting of the Association for Computational Linguistics}, \bibinfo{publisher}{Association for Computational Linguistics}, \bibinfo{address}{Online}, \bibinfo{year}{2020}, pp. \bibinfo{pages}{7871--7880}. \URLprefix \url{https://aclanthology.org/2020.acl-main.703}. \DOIprefix\doi{10.18653/v1/2020.acl-main.703}.
%Type = Inproceedings
\bibitem[{Lu et~al.(2022)Lu, Dou, and Nguyen}]{clinicalt5}
\bibinfo{author}{Q.~Lu}, \bibinfo{author}{D.~Dou}, \bibinfo{author}{T.~Nguyen},
\newblock \bibinfo{title}{{C}linical{T}5: A generative language model for clinical text},
\newblock in: \bibinfo{editor}{Y.~Goldberg}, \bibinfo{editor}{Z.~Kozareva}, \bibinfo{editor}{Y.~Zhang} (Eds.), \bibinfo{booktitle}{Findings of the Association for Computational Linguistics: EMNLP 2022}, \bibinfo{publisher}{Association for Computational Linguistics}, \bibinfo{address}{Abu Dhabi, United Arab Emirates}, \bibinfo{year}{2022}, pp. \bibinfo{pages}{5436--5443}. \URLprefix \url{https://aclanthology.org/2022.findings-emnlp.398}. \DOIprefix\doi{10.18653/v1/2022.findings-emnlp.398}.
%Type = Article
\bibitem[{Raffel et~al.(2020)Raffel, Shazeer, Roberts, Lee, Narang, Matena, Zhou, Li, and Liu}]{t5}
\bibinfo{author}{C.~Raffel}, \bibinfo{author}{N.~Shazeer}, \bibinfo{author}{A.~Roberts}, \bibinfo{author}{K.~Lee}, \bibinfo{author}{S.~Narang}, \bibinfo{author}{M.~Matena}, \bibinfo{author}{Y.~Zhou}, \bibinfo{author}{W.~Li}, \bibinfo{author}{P.~J. Liu},
\newblock \bibinfo{title}{Exploring the limits of transfer learning with a unified text-to-text transformer},
\newblock \bibinfo{journal}{Journal of Machine Learning Research} \bibinfo{volume}{21} (\bibinfo{year}{2020}) \bibinfo{pages}{1--67}.
%Type = Article
\bibitem[{Kingma and Ba(2015)}]{kingma2015adam}
\bibinfo{author}{D.~P. Kingma}, \bibinfo{author}{J.~Ba},
\newblock \bibinfo{title}{Adam: A method for stochastic optimization},
\newblock \bibinfo{journal}{arXiv preprint arXiv:1412.6980}  (\bibinfo{year}{2015}).
%Type = Inproceedings
\bibitem[{Micikevicius et~al.(2018)Micikevicius, Narang, Alben, Diamos, Elsen, Garcia et~al.}]{micikevicius2018mixed}
\bibinfo{author}{P.~Micikevicius}, \bibinfo{author}{S.~Narang}, \bibinfo{author}{J.~Alben}, \bibinfo{author}{G.~Diamos}, \bibinfo{author}{E.~Elsen}, \bibinfo{author}{D.~Garcia}, et~al.,
\newblock \bibinfo{title}{Mixed precision training},
\newblock in: \bibinfo{booktitle}{International Conference on Learning Representations}, \bibinfo{year}{2018}.
%Type = Article
\bibitem[{Sokolova and Lapalme(2009)}]{sokolova2009systematic}
\bibinfo{author}{M.~Sokolova}, \bibinfo{author}{G.~Lapalme},
\newblock \bibinfo{title}{A systematic analysis of performance measures for classification tasks},
\newblock \bibinfo{journal}{Information Processing \& Management} \bibinfo{volume}{45} (\bibinfo{year}{2009}) \bibinfo{pages}{427--437}.
%Type = Article
\bibitem[{Zhang et~al.(2019)Zhang, Kishore, Wu, Weinberger, and Artzi}]{zhang2019bertscore}
\bibinfo{author}{T.~Zhang}, \bibinfo{author}{V.~Kishore}, \bibinfo{author}{F.~Wu}, \bibinfo{author}{K.~Q. Weinberger}, \bibinfo{author}{Y.~Artzi},
\newblock \bibinfo{title}{Bertscore: Evaluating text generation with bert},
\newblock \bibinfo{journal}{arXiv preprint arXiv:1904.09675}  (\bibinfo{year}{2019}).
%Type = Inproceedings
\bibitem[{Papineni et~al.(2002)Papineni, Roukos, Ward, and Zhu}]{papineni2002bleu}
\bibinfo{author}{K.~Papineni}, \bibinfo{author}{S.~Roukos}, \bibinfo{author}{T.~Ward}, \bibinfo{author}{W.-J. Zhu},
\newblock \bibinfo{title}{Bleu: a method for automatic evaluation of machine translation},
\newblock in: \bibinfo{booktitle}{Proceedings of the 40th annual meeting of the Association for Computational Linguistics}, \bibinfo{year}{2002}, pp. \bibinfo{pages}{311--318}.
%Type = Inproceedings
\bibitem[{Lin(2004)}]{lin2004rouge}
\bibinfo{author}{C.-Y. Lin},
\newblock \bibinfo{title}{Rouge: A package for automatic evaluation of summaries},
\newblock in: \bibinfo{booktitle}{Text summarization branches out: Proceedings of the ACL-04 workshop}, \bibinfo{year}{2004}, pp. \bibinfo{pages}{74--81}.
%Type = Inproceedings
\bibitem[{Banerjee and Lavie(2005)}]{banerjee2005meteor}
\bibinfo{author}{S.~Banerjee}, \bibinfo{author}{A.~Lavie},
\newblock \bibinfo{title}{Meteor: An automatic metric for mt evaluation with improved correlation with human judgments},
\newblock in: \bibinfo{booktitle}{Proceedings of the ACL workshop on intrinsic and extrinsic evaluation measures for machine translation and/or summarization}, \bibinfo{year}{2005}, pp. \bibinfo{pages}{65--72}.
%Type = Article
\bibitem[{Brown et~al.(2020)Brown, Mann, Ryder, Subbiah, Kaplan, Dhariwal, Neelakantan, Shyam, Sastry, Askell et~al.}]{Brown}
\bibinfo{author}{T.~B. Brown}, \bibinfo{author}{B.~Mann}, \bibinfo{author}{N.~Ryder}, \bibinfo{author}{M.~Subbiah}, \bibinfo{author}{J.~D. Kaplan}, \bibinfo{author}{P.~Dhariwal}, \bibinfo{author}{A.~Neelakantan}, \bibinfo{author}{P.~Shyam}, \bibinfo{author}{G.~Sastry}, \bibinfo{author}{A.~Askell}, et~al.,
\newblock \bibinfo{title}{Language models are few-shot learners},
\newblock \bibinfo{journal}{Advances in neural information processing systems} \bibinfo{volume}{33} (\bibinfo{year}{2020}) \bibinfo{pages}{1877--1901}.
%Type = Article
\bibitem[{Luo et~al.(2022)Luo, Sun, Cheng, Zhang, Xu, Li, Zhang, Bi, Zhao, Wang et~al.}]{Lou}
\bibinfo{author}{R.~Luo}, \bibinfo{author}{K.~Sun}, \bibinfo{author}{Y.~Cheng}, \bibinfo{author}{Y.~Zhang}, \bibinfo{author}{Y.~Xu}, \bibinfo{author}{Y.~Li}, \bibinfo{author}{N.~Zhang}, \bibinfo{author}{B.~Bi}, \bibinfo{author}{X.~Zhao}, \bibinfo{author}{H.~Wang}, et~al.,
\newblock \bibinfo{title}{Biogpt: generative pre-trained transformer for biomedical text generation and mining},
\newblock \bibinfo{journal}{Briefings in Bioinformatics}  (\bibinfo{year}{2022}).

\end{thebibliography}

\appendix

\end{document}